\documentclass[10pt,twocolumn,letterpaper]{article}

\usepackage{cvpr}              

\usepackage[dvipsnames]{xcolor}

\definecolor{cvprblue}{rgb}{0.21,0.49,0.74}
\usepackage[pagebackref,breaklinks,colorlinks,citecolor=cvprblue]{hyperref}

\def\paperID{42} 
\def\confName{CVPR}
\def\confYear{2024}

\title{BaboonLand Dataset: Tracking Primates in the Wild and Automating Behaviour Recognition from Drone Videos}

\author{Isla Duporge \thanks{\texttt{Contributed equally.}}\\
{\tt\scriptsize Isla.duporge@princeton.edu}
\and
Maksim Kholiavchenko $^{*}$\\
{\tt\scriptsize kholim@rpi.edu}
\and
Roi Harel\\
{\tt\scriptsize rharel@ab.mpg.de}
\and
Scott Wolf\\
{\tt\scriptsize swwolf@princeton.edu}
\and
Dan Rubenstein\\
{\tt\scriptsize dir@princeton.edu}
\and
Meg Crofoot\\
{\tt\scriptsize mcrofoot@ab.mpg.de}
\and
Tanya Berger-Wolf\\
{\tt\scriptsize berger-wolf.1@osu.edu}
\and
Stephen Lee\\
{\tt\scriptsize stephen.j.lee28.civ@army.mil}
\and
Julie Barreau\\
{\tt\scriptsize julie.barreau@etudiant.univ-rennes.fr}
\and
Jenna Kline\\
{\tt\scriptsize kline.377@osu.edu}
\and
Michelle Ramirez\\
{\tt\scriptsize ramirez.528@.osu.edu}
\and
Charles Stewart \\
{\tt\scriptsize stewart@rpi.edu}
}

\begin{document}
\maketitle
\begin{abstract}
Using unmanned aerial vehicles (UAVs) to track multiple individuals simultaneously in their natural environment is a powerful approach for better understanding group primate behavior. Previous studies have demonstrated that it is possible to automate the classification of primate behavior from video data, but these studies have been carried out in captivity or from ground-based cameras. However, to understand group behavior and the self-organization of a collective, the whole troop needs to be seen at a scale where behavior can be seen in relation to the natural environment in which ecological decisions are made. This study presents a novel dataset for baboon detection, tracking, and behavior recognition from drone videos. The foundation of our dataset is videos from drones flying over the Mpala Research Centre in Kenya. The baboon detection dataset was created by manually annotating all baboons in drone videos with bounding boxes. A tiling method was subsequently applied to create a pyramid of images at various scales from the original 5.3K resolution images, resulting in approximately 30K images used for baboon detection. The baboon tracking dataset is derived from the baboon detection dataset, where all bounding boxes are consistently assigned the same ID throughout the video. This process resulted in half an hour of very dense tracking data. The baboon behavior recognition dataset was generated by converting tracks into mini-scenes, a video subregion centered on each animal, each mini-scene was manually annotated with 12 distinct behavior types, and one additional category for occlusion, resulting in over 20 hours of data. Benchmark results show mean average precision (mAP) of 92.62\% for the YOLOv8-X detection model, multiple object tracking precision (MOTA) of 63.81\% for the BotSort tracking algorithm, and micro top-1 accuracy of 63.97\% for the X3D behavior recognition model. Using deep learning to rapidly and accurately classify wildlife behavior from drone footage facilitates non-invasive data collection on behavior enabling the behavior of a whole group to be systematically and accurately recorded. The dataset can be accessed at \href{https://dirtmaxim.github.io/BaboonLand}{https://baboonland.xyz}.
\end{abstract}    
\section{Introduction}
\label{sec:introduction}

\begin{figure*}
  \centering
   \includegraphics[width=\linewidth]{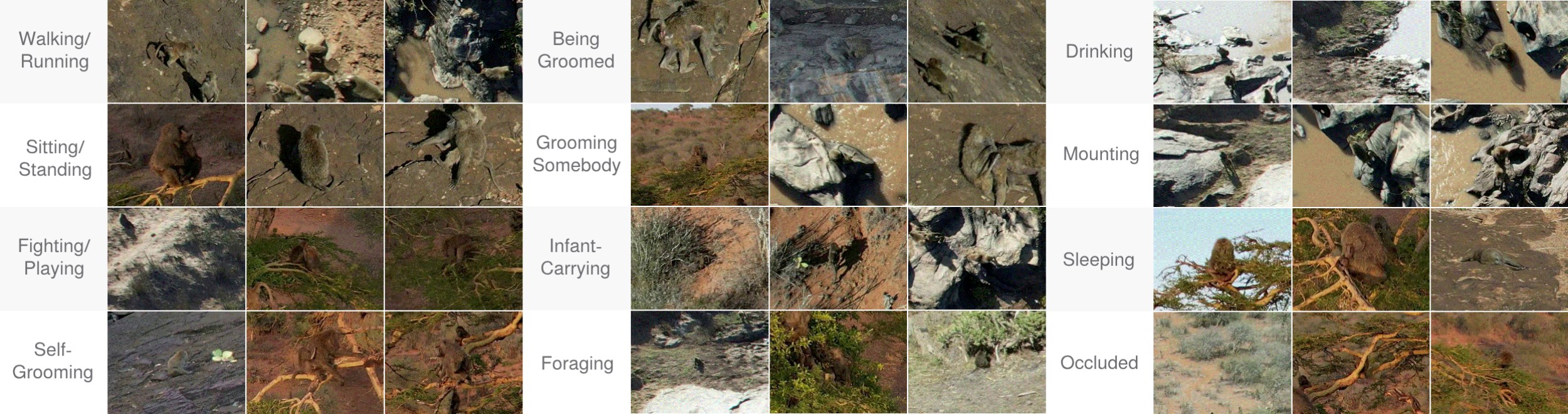}
  \caption{Examples of the behavior categories in the BaboonLand dataset: ``Walking/Running", ``Sitting/Standing", ``Fighting/Playing", ``Self-Grooming", ``Being Groomed", ``Grooming Somebody", ``Mutual Grooming", ``Infant-Carrying", ``Foraging", ``Drinking", ``Mounting", ``Sleeping", and ``Occluded".}
  \label{fig:examples_of_behaviors}
\end{figure*}

Direct observation and scoring of animal behavior has been the dominant approach in Ethology for decades~\cite{altmann1974observational}. However, manual observation is constrained by the limits of human perception, allowing only a few individuals to be monitored at once. We are in the golden age of computational ethology~\cite{anderson2014toward, kays2015terrestrial}, enabling the collection of very large datasets complemented by the possibility of automated behavior classification. Computer-based classification increases the accuracy and consistency of annotations compared to manual observers ~\cite{duporge2021using}. Several convolutional neural network (CNN)-based action recognition and pose estimation architectures have been used on animal behavior datasets ~\cite{perez2023cnn, lauer2022multi, pereira2022sleap, walter2021trex, graving2019deepposekit}, enabling insight into behavior at unprecedented levels of detail ~\cite{nath2019using}. Behavior classification software is widely used in laboratory settings on captive animals in biomedical studies and neuroscience ~\cite{mathis2020deep} e.g., recording and manipulating neural activity while measuring changes in behavior, ~\cite{bohnslav2021deepethogram, segalin2021mouse, pereira2020quantifying}; automated tracking has also been widely used to evaluate welfare and housing conditions of farm animals ~\cite{shao2021pig} and captive wild animals ~\cite{lei2022postural}. The laboratory context is controlled, and cameras are fixed, covering relatively small areas. Tracking animals in the wild is far more challenging as animals move between diverse backgrounds, with varying weather conditions, light conditions, levels of occlusion, and motion blur ~\cite{bain2021automated}. To track animal behavior in a natural context, mobile monitoring is required to capture group dynamics as animals move across a landscape throughout the day. Mobile tracking has been made possible by advances in unmanned aerial vehicles (UAVs) when flown at altitudes which mitigate visual or acoustic disturbance to wildlife ~\cite{duporge2021determination} enabling collective behavior to be observed in natural contexts ~\cite{koger2023quantifying}. This tracking method is less invasive than GPS collars and tags as the animals are not captured or tranquilized \cite{isbell2019capture}.

The study makes several contributions to the field of non-human primate tracking and action recognition:

\begin{enumerate}
\item We present a tracking and behavior recognition dataset of three troops of baboons filmed in their natural environment by drone. The dataset comprises over 30K images and annotations for baboon detection, half an hour of very dense tracking data (up to 70 individuals in a frame) and 20 hours of annotated video data of baboon behavior containing 12 distinct behavior types.
\item The BaboonLand dataset features extended behavior observations, with videos of up to 3 minutes derived from 30 minutes of continuous tracking. This is atypical for action recognition datasets, which are often compiled by scraping shorter clips from the internet. The longer videos in our dataset could be particularly useful for benchmarking transformers, known for their ability to maintain temporal context.
\item We present a benchmark for the BaboonLand dataset, addressing three key challenges: detection, tracking, and behavior recognition. Our detection model, YOLOv8-X, achieved a mean average precision (mAP) of 92.62\%. For tracking, the BotSort algorithm attained a multiple object tracking accuracy (MOTA) of 63.81\%. In behavior recognition, the X3D model achieved a micro top-1 accuracy of 63.97\%. These results reflect the complexity of the BaboonLand dataset, which comprises a dense number of small objects. The dataset provides a foundation for future research.
\end{enumerate}

Accurately detecting and tracking the social interactions of wild primates is valuable for several reasons. More accurate data on the behavior of wild populations enables insight into primate ecology and evolution, and tracking an entire troop continuously enables collective organization dynamics and social hierarchies to be understood. Undisturbed observations on social interactions can inform models of interaction dynamics, which can be used to understand the spread of disease among group mates where zoonotic spread begins \cite{Plowright2024}. Additionally, accurately tracking the proximity of primate groups to human settlement areas can provide a tool to support programs on human-primate coexistence \cite{Mitani2024}.

\section{Related Work}
\label{sec:related_work}

Several comprehensive reviews have been published on automating observations of wildlife behavior~\cite{anderson2014toward, perez2023cnn, mathis2020deep, ravoor2020deep, kleanthous2022survey}, and several multi-purpose multi-species trackers have provided vital benchmarks for the field of wildlife behavior automation, including AnimalTrack ~\cite{zhang2023animaltrack}, MammalNet~\cite{chen2023mammalnet}, and AnimalKingdom~\cite{ng2022animal}. However, these models rely on video collected from stationary ground-based sensors, including just a few individuals, limiting the ecological inferences that can be made on group behavior.  Ground-based sensors allow animals to be observed in limited spatial and temporal windows and do not provide information on how animals make ecological decisions in the wild. UAVs provide the opportunity to collect data on group behavior using an aerial perspective - for a comprehensive review on automating the identification of animals from UAV, see ~\cite{corcoran2021automated}. Several studies have used UAVs to infer social behavior but have not automatically classified behavior  ~\cite{torney2018inferring, strandburg2017habitat, maeda2021aerial}.  

Action recognition and pose estimation are two distinct tasks in computer vision. Action recognition focuses on identifying the type of action being performed (e.g., walking, running)~\cite{kholiavchenko2024kabr}, providing a label or class for the observed activity, while pose estimation determines the precise positions of a person's joints (keypoints), resulting in a set of coordinates for body parts. While action recognition emphasizes classifying the overall action, pose estimation concentrates on mapping the spatial configuration of the body. These approaches complement one another as pose estimation can enhance action recognition by providing more detailed information on body movement~\cite{luvizon20182d3d}. In this paper, we only use action recognition as up to 70 individuals are in a frame at once, meaning individual baboons appear very small, so it is not feasible to accurately identify key points for every individual. However, pose estimation is possible in ground-based videos where individuals are much larger, as demonstrated for chimpanzees ~\cite{ma2024chimpact} and macaques ~\cite{labuguen2021macaquepose, liu2022monkeytrail}. One study used hand-held video footage to achieve high accuracy in pose estimation for chimpanzees and bonobos ~\cite{wiltshire2023deepwild} and another automated recognition of two specific behaviors - buttress drumming and nut cracking in wild chimpanzees using a stationary video camera and audio to locate times in the video when these behaviors occurred ~\cite{bain2021automated}. 

Applying behavior recognition to videos with multiple individuals in the frame requires first localizing each individual. Some approaches utilize end-to-end models to simultaneously predict the localized actions of all individuals in the video ~\cite{li2020ava}. However, this approach is not feasible in our case as resizing the original 5.3K video frames down to a size the CNN can handle results in an image where all individuals are scaled down to such a low resolution that their actions cannot be distinguished. In our paper, we utilize the mini-scenes approach initially introduced by KABR~\cite{kholiavchenko2024kabr}. The authors demonstrated that using mini-scenes is an efficient approach to solving animal behavior recognition problems when the animal is very small compared to the size of the videos. 

The most relevant dataset to BaboonLand is PanAf20K - a video dataset of Wild Apes collected from camera traps ~\cite{brookes2024panaf20k}; however, this data is from stationary cameras containing very few individuals in the frame at any one time and captured from a fixed geographical position; thus this cannot be used to observe collective group behavior.  BaboonLand demonstrates the ability to combine advances in UAV technology with action recognition to automate ethograms of primate behavior, tracking behavior as it occurs for large groups of individuals in natural habitats as they move across a landscape. The dataset is diverse as it covers several contrasting environmental backgrounds, i.e., a cliff, river crossing, open savannah, rock, and sleeping tree; it also includes a very dense number of individuals and automatically classifies the behavior of all individuals in the group. Examples of the behavior categories are depicted in \cref{fig:examples_of_behaviors}. The behavior of the collective can be observed in its natural environment, which provides the opportunity to ask important biological questions about individual behavior in relation to other individuals in the troop, enabling dyadic behaviors to be observed and the dynamics of social hierarchies to be deciphered. 

A challenge in tracking wild animals is maintaining individual identities when numerous animals move in and out of frame or close together ~\cite{agezo2022tracking}. This is less problematic for species with unique coat signatures, as deep learning-assisted identification uses these cues to assign IDs ~\cite{nepovinnykh2024norppa, bergler2021fin}. Still, it remains a challenge for individuals without distinct markings ~\cite{ravoor2020deep}. In the laboratory, temporary distinct markings can be used e.g., painting ear tufts ~\cite{lauer2022multi} or attaching QR codes ~\cite{wolf2023naps}. However, this is not possible in the wild as it is logistically challenging and interferes with behavior. Face recognition models have been used to assign individual identities for several primates~\cite{brust2017towards}, ~\cite{witham2018automated}, ~\cite{deb2018face}, \cite{crouse2017lemurfaceid}, ~\cite{schofield2019chimpanzee, loos2015face} however, detectors fail when faces are obscured from view ~\cite{shukla2019primate}, which is frequently the case when tracking from an aerial perspective. Several methods have been developed to track unmarked individuals but have yet to be demonstrated on large numbers of individuals in the wild \cite{Romero-Ferrero2019}.
\section{The BaboonLand Dataset}

Olive baboons (\textit{Papio anubis}) were used as the test species for this study; however, the model can be applied to other non-human primates on retraining. Olive Baboons have a well-described catalog of behaviors and have historically been used as a model for human evolution ~\cite{strum2012darwin}; they live in relatively stable, cohesive groups of 15-100 individuals ~\cite{altmann1970baboon} and exhibit rich behavioral repertoires. They are the most extensively distributed of the baboons, ranging across forest, savannah, shrubland, grassland, and urban areas in East, West, and Southern Africa ~\cite{wallis2020papio}. They have two main predators - leopards and lions ~\cite{cowlishaw1994vulnerability}, which they avoid by returning to sleeping sites selected as safe ~\cite{strandburg2017habitat}. We filmed three troops of baboons as they left and returned to their sleeping sites at three known locations, including a sleeping tree by a river camp and two sleeping cliffs. The troops of baboons reside in Mpala, an open landscape in Laikipia county, central Kenya, comprising savannah and dry woodland habitat bordered by the Ewaso Ngiro and Ewaso Narok rivers. Mpala is not fenced; thus, these wild primates were observed in an unconstrained natural environment.

All video data was collected via a DJI Air 2S with a 1-inch image sensor, 20MP; 2.4\textmu m pixel size with videos shot in 5.3K. Test flights were conducted to check disturbance altitude; the baboons showed no disturbance when flights were conducted above 15 meters. To add redundancy, all flights were conducted above 20 meters altitude. The troops were filmed as they left and returned from their known sleeping sites, morning and night. Troops were followed for as long as possible before either total battery life was depleted or we lost track of the troop on the landscape. The videos were collected in various background environments, including when sleeping on a tree, on a rock, during a river crossing, in an open savannah, and on a cliff. The data contains lots of visual ``noise" from camera motion, varying light conditions, spectral contrast from shadow, and different background environments. These variable conditions are beneficial for training robust models, as they promote greater adaptability and resilience.

\begin{figure}[t]
  \centering
   \includegraphics[width=1.0\linewidth]{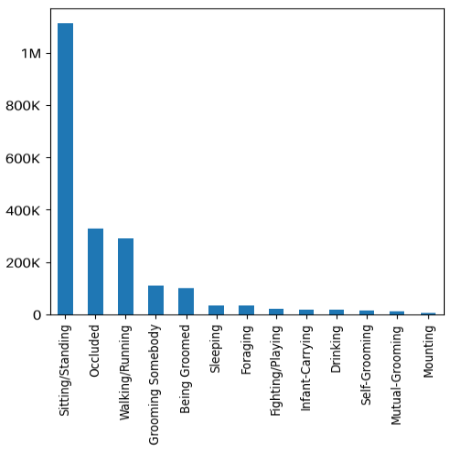}
  \caption{The distribution of behavior classes in the BaboonLand dataset.}
  \label{fig:stat}
\end{figure}

The baboon detection dataset was created by manually annotating all baboons in the videos with bounding boxes. A tiling method was then used to generate a pyramid of images at different scales from the original 5.3K resolution footage, resulting in approximately 30K images for baboon detection. The baboon tracking dataset was derived from the detection dataset, with each bounding box consistently assigned the same ID throughout the video, yielding half an hour of dense tracking data. The tracks were converted into mini-scenes for the baboon behavior recognition dataset, which are video subregions centered on each animal. Each mini-scene was manually annotated with 12 distinct behavior types and an additional category for occlusions, resulting in about 20 hours of data. Our behavior dataset exhibits a long-tailed distribution, indicating significant disparities in the number of samples across different categories. This is anticipated, as some behaviors are naturally more prevalent in baboons than others. The class distribution is illustrated in \cref{fig:stat}.

We provide the original drone videos along with the corresponding tracks and bounding boxes. Additionally, the dataset includes several evaluation sets: one for tracking algorithms (with 75\% of each video used for training and 25\% for testing), a YOLO-formatted dataset for training and evaluating detection (80\% of images for training, 7\% for validation, and 13\% for testing) ~\cite{Jocher_Ultralytics_YOLO_2023}, and a dataset in the Charades format for training and evaluating behavior recognition models (75\% of videos for training and 25\% for testing) ~\cite{sigurdsson2016hollywood}. We also provide all of the data processing scripts that we used to generate these datasets from original drone videos. Tracks and behavior annotations in the BaboonLand dataset are stored in the simplified CVAT for video 1.1 format. These can be uploaded to CVAT for exploratory data analysis and annotation adjustments. After making any annotation adjustments, the provided scripts can be used to regenerate all of the sub-datasets (detection, tracking, and behavior recognition).
\section{Annotation and Review}

All videos from the drone flights were reviewed, and 18 were selected based on the diversity of their backgrounds and baboon activity. Computer Vision Annotation Tool (CVAT) was used to manually annotate bounding boxes, tracks, and individual behaviors. As manual tracking and behavioral scoring are subject to human bias, nine annotators participated in the labeling process to minimize this bias. Then a separate annotator went through the video to check all annotations. For behavior annotation, each video was also checked by two separate annotators to ensure behavior classes were agreed on, and a third annotator then checked a random sampling of these videos.
\section{Experiments}

\begin{figure}[t]
  \centering
   \includegraphics[width=1.0\linewidth]{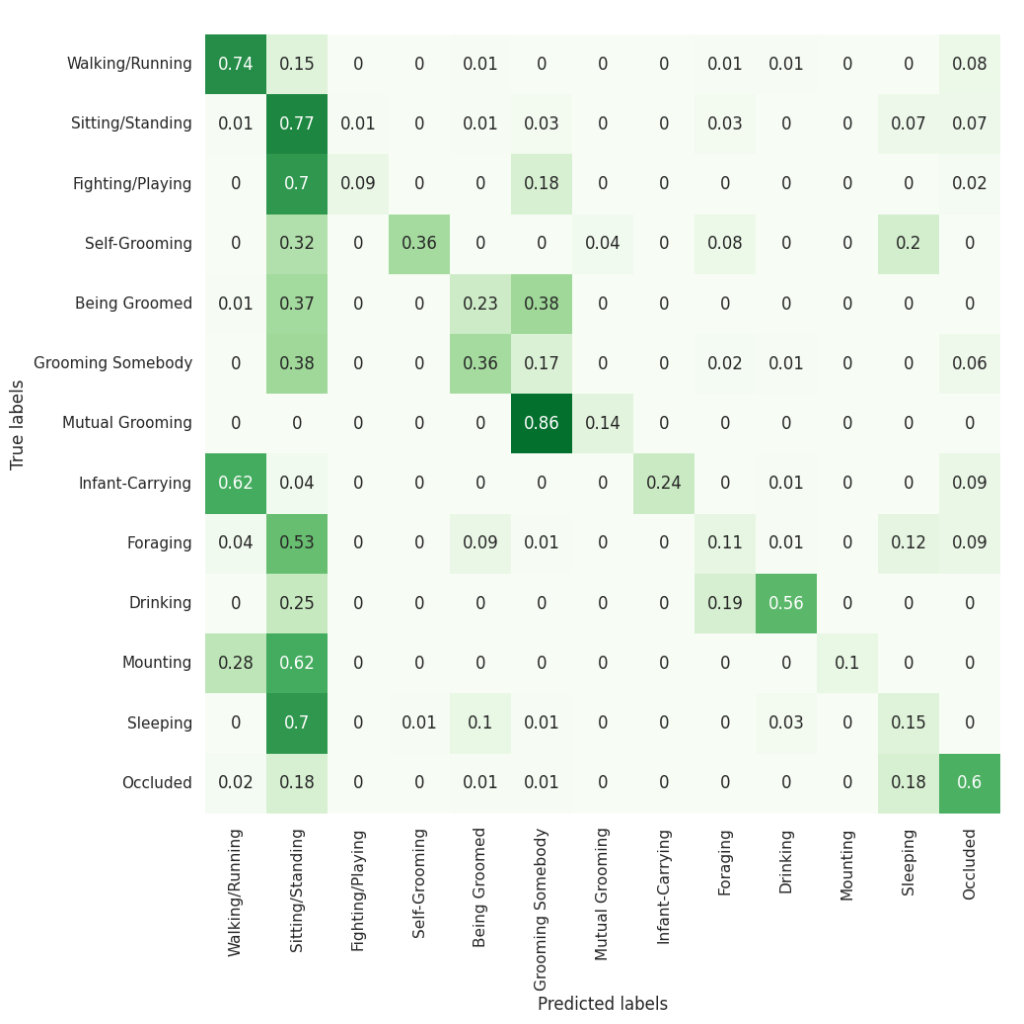}
  \caption{This confusion matrix showcases the performance of the X3D model, which is the top-performing model on the BaboonLand dataset based on our evaluation.}
  \label{fig:cm}
\end{figure}

\subsection{Detection}
Our dataset for object detection was evaluated using YOLOv8-X~\cite{Jocher_Ultralytics_YOLO_2023}.
Given the small size of the baboons in the original footage, a specialized approach was necessary to train a detector effectively. We adopted a tiled methodology, generating a multi-scale pyramid (2x2, 3x3, 4x4) to encompass varying perspectives and dimensions of the target object.

\begin{table}[h]
  \centering
  \begin{tabular}{p{1.7cm}p{1.65cm}p{1.65cm}p{1.65cm}}
    \toprule
    \hfil Model & \hfil mAP@50 & \hfil Precision & \hfil Recall \\
    \midrule
    \hfil YOLOv8-X & \hfil \textbf{92.62} & \hfil \textbf{93.70} & \hfil \textbf{87.60} \\
    \bottomrule
  \end{tabular}
  \caption{The results of YOLOv8-X model trained on our dataset. The model was trained for 64 epochs with 768x768 input resolution.}
  \label{tab:example}
\end{table}

\subsection{Tracking}

We adopted ByteTrack~\cite{zhang2022bytetrack} and BotSort~\cite{aharon2022bot} algorithms to evaluate the tracking performance. Both algorithms were built on the trained YOLOv8-X model described in the preceding section. The effectiveness of the tracking algorithm depends on several factors, including the quality of baboon detection and the precision of the detection association. In our evaluation, BotSort demonstrated superior results compared to ByteTrack.

\begin{table}[h]
  \centering
  \begin{tabular}{cccccc}
    \toprule
    Tracker & MOTA & MOTP & IDF1 & P & R\\
    \midrule
    ByteTrack & 63.55 & 34.10 & 77.01 & 96.32 & 64.90 \\
    BotSort & \textbf{63.81} & \textbf{34.31} & \textbf{78.24} & \textbf{97.21} & \textbf{66.16} \\
    \bottomrule
  \end{tabular}
  \caption{Evaluation results of ByteTrack and BotSort. The YOLOv8-X model was a backbone for both ByteTrack and BotSort algorithms.}
  \label{tab:example}
\end{table}

\subsection{Behavior Recognition}

To provide the baseline for behavior recognition, we trained I3D~\cite{carreira2017quo}, SlowFast~\cite{feichtenhofer2019slowfast}, and X3D~\cite{feichtenhofer2020x3d} models on our dataset. We report micro (per instance) average and macro (per class) average accuracy. The confusion matrix depicted in \cref{fig:cm} demonstrates
the performance of the X3D model. We can see that the model performs quite well for common classes but rare behaviors are more challenging. The model tends to predict ``Sitting/Standing for actions that encompass similar behaviors, such as ``Drinking," ``Foraging," and ``Mounting". This indicates that while the model is effective at identifying frequent activities, it has difficulty distinguishing between less common behaviors that share visual similarities.

\begin{table}[h]
  \centering
  \begin{tabular}{ccp{1.2cm}p{1.2cm}p{1.2cm}}
    \toprule
    Average & Method & Top-1 & Top-3 &  Top-5 \\
    \midrule
    & I3D & 26.53 & 54.51 & 65.47 \\
    Macro & SlowFast & 27.08 & 56.73 & 67.61 \\
    & X3D & \textbf{30.04} & \textbf{60.58} & \textbf{72.13} \\
    \midrule
    & I3D & 61.29 & 89.38 & 92.34 \\
    Micro & SlowFast & 61.71 & 90.35 & 93.11 \\
    & X3D & \textbf{63.97} & \textbf{91.34} & \textbf{95.17} \\
    \bottomrule
  \end{tabular}
  \caption{Results of I3D, SlowFast, and X3D models. I3D and X3D were trained with 16 input frames with a sampling rate of 5. For SlowFast, the Slow branch was trained with 16 input frames with a sampling rate of 5, and the Fast branch was trained with 4 input frames with a sampling rate of 5.}
  \label{tab:example_micro}
\end{table}
\section{Ethical Considerations}

No humans can be distinguished in the videos, and the research was conducted under the authority of a Nacosti Research License. This license confirms adherence to the regulations enabling drone footage of animals to be collected in their natural habitats. We followed a protocol that strictly complies with the guidelines set forth by the Institutional Animal Care and Use Committee (IACUC). These guidelines are designed to ensure the ethical and humane treatment of animals involved in research activities. We flew at an altitude that did not disturb the baboons after calibrating this via several trial flights. We approached the animals from downwind, allowing drone noise to dissipate before reaching the animals.
\section{Conclusion}

This paper presents a unique dataset encompassing a complex range of drone videos of Olive baboons moving between various background contexts, it can be used as a dataset to evaluate detection, tracking, and behavior recognition models. Our experiments demonstrate that the proposed dataset is a challenge for the evaluation of new state-of-the-art algorithms. The study of behavioral transitions of baboons provides a promising direction to explore how individuals are affected by their social environment, examining dyadic and ultra-dyadic interactions. The method we present here which automatically tracks collective behavior from UAV will be applicable to satellite video footage in the future. It is already possible to directly detect animal groups in satellite imagery \cite{duporge2022satellite,duporge2021using} and several companies are now developing satellites that can collect video. This will provide a more sophisticated method to monitor wildlife at grander spatial scales in the future.
\section{Acknowledgments}

This material is based upon work supported by the National Science Foundation under Award No. 2118240 and Award No. 2112606. ID was supported by the National Academy of Sciences Research Associate Program and the United States Army Research Laboratory while conducting this study. ID collected all the UAV data on a Civil Aviation Authority Drone License CAA NQE Approval Number: 0216/1365 in conjunction with authorization from a KCAA operator under a Remote Pilot License. The data was gathered at the Mpala Research Centre in Kenya, in accordance with Research License No. NACOSTI/P/22/18214. The data collection protocol adhered strictly to the guidelines set forth by the Institutional Animal Care and Use Committee under permission No. IACUC 1835F.
{
    \small
    \bibliographystyle{unsrt}
    \bibliography{main}
}


\end{document}